\documentclass[twoside,11pt]{article}

\usepackage{bm}
\usepackage{jmlr2e}

\usepackage{booktabs}
\usepackage{listings}
\usepackage{comment}
\usepackage{lineno,hyperref}
\usepackage{amsmath}
\usepackage{verbatim}
\usepackage[ruled,vlined]{algorithm2e}

\usepackage{float}
\usepackage{xfrac}
\usepackage[all]{xy}
\usepackage{siunitx}
\usepackage{color}
\usepackage{nth}
\usepackage{pgfplots}
\pgfplotsset{compat=1.14}

\pgfplotsset{compat=newest}
\pgfplotsset{plot coordinates/math parser=false}
\usepackage{tikzscale}
\usetikzlibrary{matrix,chains,positioning,decorations.pathreplacing,arrows}
\usepackage{tikz-qtree,tikz-qtree-compat}
\usetikzlibrary{calc}

\ShortHeadings{Explainable AI for the Medical Domain}{Holzinger, Biemann, Pattichis, Kell}
\firstpageno{1}

\newcommand{\x}{\boldsymbol{x}}
\newcommand{\w}{\boldsymbol{w}}
\newcommand{\z}{\boldsymbol{z}}

\begin{document}

\title{What do we need to build explainable AI systems\\
for the medical domain?}

\author{\name Andreas Holzinger\textsuperscript{1} \email a.holzinger@hci-kdd.org\\
\name Chris Biemann\textsuperscript{2}  \email biemann@informatik.uni-hamburg.de\\
\name Constantinos S. Pattichis\textsuperscript{3} \email pattichi@cs.ucy.ac.cy\\
\name Douglas B. Kell\textsuperscript{4}\email dbk@manchester.ac.uk\\
\\
\addr \textsuperscript{1} Holzinger Group HCI-KDD, Inst. for Med. Informatics/Statistics,\,Medical\,University\,Graz,\,Austria\\
\addr \textsuperscript{2} Language Technology Group LT, Department of Informatics, University of Hamburg, Germany\\
\addr \textsuperscript{3} E-Health Laboratory, Department of Computer Science, University of Cyprus, Cyprus\\
\addr \textsuperscript{4} School of Chemistry and Manchester Institute of Biotechnology, University of Manchester, UK\\
}


\maketitle

\begin{abstract}
Artificial intelligence (AI) generally and machine learning (ML) specifically demonstrate impressive practical success in many different application domains, e.g. in autonomous driving, speech recognition, or recommender systems. Deep learning approaches, trained on extremely large data sets or using reinforcement learning methods have even exceeded human performance in visual tasks, particularly on playing games such as Atari, or mastering the game of Go. Even in the medical domain there are remarkable results. However, the central problem of such models is that they are regarded as black-box models and even if we understand the underlying mathematical principles of such models they lack an explicit declarative knowledge representation, hence have difficulty in generating the underlying explanatory structures. This calls for systems enabling to make decisions transparent, understandable and explainable. A huge motivation for our approach are rising legal and privacy aspects. The new European General Data Protection Regulation (GDPR and ISO/IEC 27001) entering into force on May 25th 2018, will make black-box approaches difficult to use in business. This does not imply a ban on automatic learning approaches or an obligation to explain everything all the time, however, there must be a \textit{possibility} to make the results re-traceable on demand. This is beneficial, e.g. for general understanding, for teaching, for learning, for research, and it can be helpful in court. In this paper we outline some of our research topics in the context of the relatively new area of explainable-AI with a focus on the application in medicine, which is a very special domain. This is due to the fact that medical professionals are working mostly with distributed heterogeneous and complex sources of data. In this paper we concentrate on three sources: images, *omics data and text. We argue that research in explainable-AI would generally help to facilitate the implementation of AI/ML in the medical domain, and specifically help to facilitate transparency and trust.
\end{abstract}


\section{Introduction and Motivation}
\label{section:introduction}

Artificial intelligence (AI) has a long tradition in computer science and experienced many ups and downs since its formal introduction as an academic discipline six decades ago \citep{Holland:1992:AdaptationBook,RusselNorvig:1995:AIbook}. The field recently gained enormous interest, mostly due to the practical successes in Machine Learning (ML). The grand goal of AI is to provide the theoretical fundamentals for ML to develop software that can learn autonomously from previous experience, automatically and with no human-in-the-loop \citep{ShahriariAdamsFreitas:2016:HumanOutofTheLoop}. Ultimately, to reach a level of \textit{usable intelligence,} we need (1) to learn from prior data, (2) to extract knowledge, (3) to generalize, (4) to fight the curse of dimensionality, and (5) to disentangle the underlying explanatory factors of the data \citep{BengioVincent:2013:RepresentationLearning}. One grand challenge still remains open: to make sense of the data in the \textbf{context} of an application domain. The quality of data and appropriate features matter most, and previous work has shown that the best-performing methods typically combine multiple low-level features with high-level context \citep{GirshickDonahueDarrellMalik:2014:Features}. However, the full effectiveness of all AI/ML success is limited by the algorithm's inabilities to explain its results to human experts - but exactly this is a big issue in the medical domain.

In the medical domain we are facing complex challenges particularly in the integration, fusion and mapping of various distributed and heterogeneous data in arbitrarily high dimensional spaces. Consequently, \textit{explainable-AI} in the context of medicine must take into account that diverse data may contribute to a relevant result. This requires that medical professionals must have a possibility \textit{to understand how and why a machine decision has been made.} Moreover, transparent algorithms could appropriately enhance trust of medical professionals in future AI systems. Research towards building explainable-AI systems for application in medicine requires to maintain a high level of learning performance for a range of machine learning and human-computer interaction techniques. There is an inherent tension between machine learning performance (predictive accuracy) and explainability. Often the best-performing methods (e.g., deep learning) are the least transparent, and the ones providing a clear explanation (e.g., decision trees) are less accurate \citep{BolognaHayashi:2017:deep}.

The performance of algorithms also depends on the choice of the data representations; hence much engineering effort goes into the design of pre-processing pipelines and in handcrafted integration, fusion, data transformations, and mappings that result in an appropriate representation. Current automatic learning algorithms still have an enormous weakness: they are unable to extract the discriminative knowledge from the data. Consequently, it is of utmost importance to expand the applicability of learning algorithms, hence, to make them less dependent on feature engineering.
As already mentioned before, ultimately, a truly intelligent system must be able to understand the context, a long awaited but still far-off goal \citep{Zadeh:2008:HumanLevelIntelligence}.

The topic of explainable-AI is of such great importance that the U.S. Defense Advanced Research Projects Agency (DARPA) has recently set an explainable-AI (XAI) program on its agenda\footnote{https://www.darpa.mil/program/explainable-artificial-intelligence} \citep{Gunning:2016:DARPA-explainable-AI}. DARPA emphasizes the importance of Human--Computer Interaction (HCI) for machine learning and knowledge discovery/data mining (KDD). This is manifested in the HCI-KDD approach, which fosters integrative ML, i.e. a synergistic combination of diverse methodological approaches in a concerted effort to augment human intelligence with artificial intelligence, and eventually to enable what neither of them could do on their own \citep{Holzinger:2012:DATAconf,Holzinger:2013:HCI-KDD,Holzinger:2017:InauguralMAKE,HolzingerGoebelPaladeFerri:2017:Integrative}.

In the following we provide an incomplete overview on some state-of-the-art of explainable models generally and selected research on explainable models for images, *omics data and text specifically. This trilogy of data combination is needed for future medicine.

\newpage 


\section{Explainability}
\label{sec:explainability}

The problem of explainability is as old as AI and maybe the result of AI itself: whilst AI approaches demonstrate impressive practical success in many different application domains, their effectiveness is still limited by their inability to "explain" their decisions in an understandable way \citep{CoreEtAl:2006:eXAIsystems}. Even if we understand the underlying mathematical theories it is complicated and often impossible to get insight into the internal working of the models and to explain how and why a result was achieved. Explainable-AI is an rapidly emerging research area with increasing visibility in the popular press\footnote{https://www.computerworld.com.au/article/617359} and even daily press\footnote{https://www.nytimes.com/2017/11/21/magazine/can-ai-be-taught-to-explain-itself.html}.

In the pioneering days of AI \citep{NewellShawSimon:1958:ChessComplexity}, the predominant reasoning methods were logical and symbolic. These early AI systems reasoned by performing some form of logical inference on human readable symbols. Interestingly, these early systems were able to provide \textit{a trace of their inference steps} and became the basis for explanation. There is some related work available on how to make such systems explainable \citep{ShortliffeBuchanan:1975:InexactReasoning,SwartoutEtAl:1991:Explanations,Johnson:1994:AgentsExplain,LacaveDiez:2002:ExplanationBayesNets}.

In the medical domain there is growing demand in AI approaches, which are not only well performing, but trustworthy, transparent, interpretable and explainable. Methods and models are necessary to reenact the machine decision-making process, to reproduce and to comprehend both the learning and knowledge extraction process. This is important, because for decision support it is necessary to understand the \textbf{causality} of learned representations \citep{Pearl:2009:Causality,GershmanHorvitzTenenbaum:2015:ComputationalRationality,PetersJanzigSchoelkopf:2017:CausalityBook}.

Understanding, interpreting, explaining are often used synonymously in the context of explainable-AI \citep{DoranBesold:2017:ExplainableAI}, and various techniques of interpretation have been applied in the past. There is a nice discussion on the "Myth of model interpretability" by \cite{Lipton:2016:MythosInterpretability}. In the context of explainable-AI the term ``understanding'' usually means a \textit{functional understanding} of the model, in contrast to a low-level algorithmic understanding of it, i.e. to seek to characterize the model's black-box behavior, without trying to elucidate its inner workings or its internal representations.
\cite{MontavonSamekMueller:2017:InterpertingDL} discriminate in their work between \textit{interpretation}, which they define as a mapping of an abstract concept into a domain that the human expert can perceive and comprehend; and \textit{explanation}, which they define as a collection of features of the interpretable domain, that have contributed to a given example to produce a decision.

We argue that in the medical domain, something like “explainable medicine” would be urgently needed for many purposes including medical education, research and clinical decision making. If medical professionals are complemented by sophisticated AI systems and in some cases even overruled, the human experts must still have a chance, on demand, to understand and to retrace the machine decision process. However, we also point out that it is often assumed that humans are able to explain their decisions. This is often \textit{not} the case; sometimes experts are not able, or even not willing to provide an explanation.

Explainable-AI calls for confidence, safety, security, privacy, ethics, fairness and trust \citep{KiesebergEtAl:2016:TrustDocInLoop}, and puts usability on the research agenda, too \cite{MillerHoweSonenberg:2017:ExplainableAI}. All these aspects together are crucial for applicability in medicine generally, and for future personalized medicine specifically \citep{HamburgCollins:2011:PersonalizedMedicine}.



\section{Explainable Models}

We can distinguish two types of explainability/interpretability, which can be denominated with Latin names used in law \citep{FellmethHorwitz:2009:LatinLaw}: post-hoc explainability = "(lat.) after this", occurring after the event in question; e.g., explaining what the model predicts in terms of what is readily interpretable; ante-hoc explainability = "(lat.) before this", occurring before the event in question; e.g., incorporating explainability directly into the structure of an AI-model, explainability by design.

\textbf{Post-hoc systems} aim to provide local explanations for a specific decision and making it reproducible on demand (instead of explaining the whole systems behaviour). A representative example is LIME (Local Interpretable Model-Agnostic Explanations) developed by \cite{RibeiroSinghGuestrin:2016:Trust}, which is a \textbf{model-agnostic} system, where $x \in \mathbb{R}^d$ is the original representation of an instance being explained, and $x' \in \mathbb{R}^{d'}$ is used to denote a vector for its interpretable representation (e.g. $x$ may be a feature vector containing word embeddings, with $x'$ being the bag of words). The goal is to identify an interpretable model over the \emph{interpretable representation} that is \textbf{locally faithful} to the classifier. The explanation model is $g: \mathbb{R}^{d'}\rightarrow\mathbb{R}, g \in G$, where $G$ is a class of potentially interpretable models, such as linear models, decision trees, or rule lists; given a model $g \in G$, it can be visualized as an explanation to the human expert (for details please refer to \citep{RibeiroSinghGuestrin:2016:ModelAgnosticInterpret}). Another example for a post-hoc system is BETA (Black Box Explanations through Transparent Approximations, a model-agnostic framework for explaining the behavior of any black-box classifier by simultaneously optimizing for fidelity to the original model and interpretability of the explanation introduced by \cite{LakkarajuLeskovec:2017:Interpretable}. \cite{BachMueller:2015:PixelWiseExplanation} presented a general solution to the problem of understanding classification decisions by pixel-wise decomposition of nonlinear classifiers which allows to visualize the contributions of single pixels to predictions for kernel-based classifiers over bag of words features and for multilayered neural networks.

\textbf{Ante-hoc systems} are interpretable by design towards glass-box approaches \citep{HolzingerEtAl:2017:glassbox}; typical examples include linear regression, decision trees and fuzzy inference systems. The latter have a long tradition and can be designed from expert knowledge or from data and provides - from the viewpoint of HCI - a good framework for the interaction between human expert knowledge and hidden knowledge in the data \citep{Guillaume:2001:InterpretableFuzzy}. A further example was presented by \cite{CaruanaEtAl:2015:AnteHocHealth}, where high-performance generalized additive models with pairwise interactions (GAMs) were applied to problems from the medical domain yielding intelligible models, which uncovered surprising patterns in the data that previously had prevented complex learned models from being fielded in this domain; important is that they demonstrated \textit{scalability} of such methods to large data sets containing hundreds of thousands of patients and thousands of attributes while remaining intelligible and providing accuracy comparable to the best (unintelligible) machine learning methods. A further example for ante-hoc methods can be seen in \cite{PoulinEtAl:2006:VisualExplanation}, where they described a framework for visually explaining the decisions of any classifier that is formulated as an additive model and showed how to implement this framework in the context of three models: na\"{\i}ve Bayes, linear support vector machines and logistic regression, which they implemented successfully into a bioinformatics application \citep{SzafronEtAl:2004:ProteomeAnalyst}.

\newpage 

\subsection{Example: Interpreting a Deep Neural Network}

Deep neural networks (DNN), particularly convolutional neural networks (CNN) and recurrent neural networks (RNN) have demonstrated to be applicable to a wide range of practical problems, from image recognition \citep{SimonyanZisserman:2014:DeepImageRecognition} and image classification \citep{EstevaThrun:2017:DermaNN} to movement recognition \citep{SinghEtAl:2017:RNN-AAL}. At the same time these approaches are also theoretically interesting, because humans organize their ideas also hierarchically \citep{BengioY:2009:deepLearning,Schmidhuber:2015:DLOverview}.

Basically, a neural network (NN) is a collection of neurons organized in a sequence of multiple layers, where neurons receive as input the neuron activations from the previous layer, and perform a simple computation (e.g. a weighted sum of the input followed by a nonlinear activation). The neurons of the network jointly implement a complex nonlinear mapping from the input to the output. This mapping is learned from the data by adapting the weights of each individual neuron using  backpropagation, which repeatedly adjusts the weights of the connections in the network in order to minimize the difference between the current output vector and the desired output vector. As a result of the weight adjustments, internal hidden units which are not part of the input or output come to represent important features of the task domain, and the regularities in the task are captured by the interactions of these units (refer to the original paper of \cite{RumelhartHintonWilliams:1986:Backpropagation} and the review by \cite{WidrowLehr:1990:backprop} for an overview).

Typically, deep neural networks are trained using supervised learning on large and carefully annotated data sets. However, the need for such data sets restricts the space of problems that can be addressed. This has led to a proliferation of deep learning results on the same tasks using the same well-known data sets \citep{RolnickEtAl:2017:DeepRobust}.

Annotated data is extremely difficult to obtain, especially for classification tasks with large numbers of classes (requiring extensive annotation) or with fine-grained classes (requiring skilled annotation by domain experts). Consequently, annotation can be very expensive and, for tasks requiring expert knowledge, may simply be unattainable at large scale - which is often a huge problem in the medical domain. Just to illustrate this problem on a popular example: the collection of ImageNet\footnote{ImageNet is an image database containing 14,197,122 images (as of 24.12.2017) organized according to nouns of WordNet, and is openly available: http://www.image-net.org} data required more than a year of human labor on Amazon Mechanical Turk. A large-scale ontology of images is a critical resource for developing advanced, large-scale content-based image search and image understanding algorithms, as well as for providing critical training and benchmarking data for such algorithms  \citep{DengLiEtAl:2009:ImageNet}.

\cite{MontavonSamekMueller:2017:InterpertingDL} provide an excellent example of the problem of interpreting a concept learned by a deep neural network. A learned concept to be interpreted can be represented by neurons in the top layer. The problem is that top-layer neurons are abstract and not perceivable by a human, but the input domain of the network (both image or text) is usually interpretable. Thus something is necessary in the input domain, which is both interpretable and representative of the abstract learned concept - one possibility is to make use of the activation maximization principle\footnote{presented as a poster during the ICML 2009 workshop on Learning Feature Hierarchies, http://www.cs.toronto.edu/~rsalakhu/deeplearning/program.html},
\citep{ErhanBengioCourvilleVincent:2009:TechnicalReportVisDeep}.


Activation maximization can be used as an analysis framework that searches for an input pattern to produce a maximum model response for a specific quantity of interest \citep{Berkes:Wiskott:2006:QuadraticForms,SimonyanZisserman:2014:DeepImageRecognition}.
Consider a neural network classifier mapping data points $\x$ to a set of classes $(\omega_c)_c$. The output neurons encode the modeled class probabilities $p(\omega_c|\x)$, and a prototype $\x^\star$ as representative of the class $\omega_c$ can be found by optimizing:

\begin{equation}
\max_{\x} ~ \log p(\omega_c|\x) - \lambda \|\x\|^2.
\end{equation}

The class probabilities modeled by the neural net are functions having a gradient, consequently a widely-used technique in AI, gradient ascent, which aims at maximizing an objective function (opposite to gradient descent which aims to minimizing an objective function) \citep{Zinkevich:2003:gradientAscent}. The term of the objective is an $\ell_2$-norm regularizer that implements a preference for inputs that are close to the origin. When applied to image classification, prototypes thus take the form of mostly gray images, with only a few edge and color patterns at strategic locations \citep{SimonyanZisserman:2014:DeepImageRecognition}.


In order to focus on higher probable regions of the input space, the $\ell_2$-norm regularizer can be replaced by a data density model $p(\x)$ which is called ``expert'' by \cite{MontavonSamekMueller:2017:InterpertingDL}. This leads to the following optimization problem:

\begin{equation}
\max_{\x} ~ \log p(\omega_c|\x) + \log p(\x).
\end{equation}

Here, the prototype is encouraged to simultaneously produce strong class response and to resemble the data. By application of Bayes' rule, the newly defined objective can be identified, up to modeling errors and a constant term, as the class-conditioned data density $p(\x|\omega_c)$. The learned prototype thus corresponds to the most likely input $\x$ for the class $\omega_c$. A possible choice for the expert is the Gaussian Restricted Boltzmann Machine (RBM). The RBM is a two-layer, bipartite, undirected graphical model with a set of binary hidden units $p(h)$, a set of (binary or real-valued) visible units $p(v)$, with symmetric connections between the two layers represented by a weight matrix $W$. The probabilistic semantics for an RBM is defined by its energy function (for details see the chapter by \cite{Hinton:2012:RBM}. Its probability function can be written as:

\begin{equation}
\log p(\x) = {\textstyle \sum_j} f_j(\x) - {\textstyle \frac12} \x^\top \Sigma^{-1} \x + \text{cst.}
\end{equation}

where

\begin{equation}
f_j(\x) = \log(1+\exp(\w_j^\top \x + b_j))
\end{equation}

are factors with parameters learned from the data.

When interpreting more complex concepts such as natural images classes, other density models such as convolutional RBM's \citep{LeeNg:2009:convolutionalDeepBelief} or pixel RNN's \citep{OordEtAl:2016:PixelRNN} are better suitable.

\begin{figure}[h!]\centering
\includegraphics[width=0.97\linewidth]{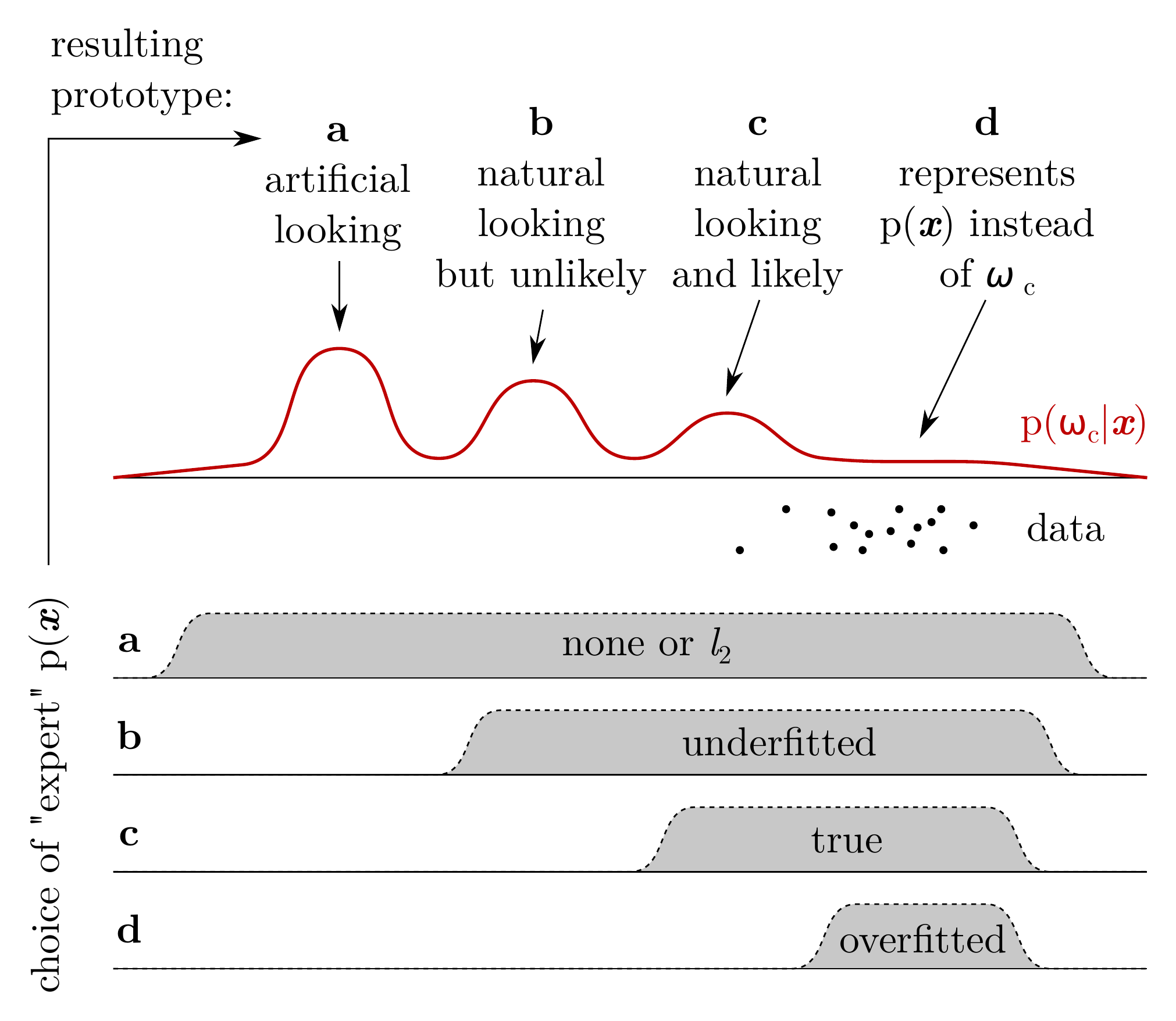}
\caption{Four cases illustrating how the "expert" $p(\x)$ affects the prototype $\x^\star$ found by activation maximization. The horizontal axis represents the input space, and the vertical axis represents the probability (extreme case a: expert is absent, extreme case d: expert is overfitted; Image source: \citep{MontavonSamekMueller:2017:InterpertingDL}.}
\label{figure:expert}
\end{figure}

The selection of the expert $p(\x)$ plays an important role. The relation between the expert and the resulting prototype is given qualitatively in Figure~\ref{figure:expert} by \cite{MontavonSamekMueller:2017:InterpertingDL}: Here we see four different cases: In case a the expert is absent, i.e. the optimization problem reduces to the maximization of the class probability function $p(\omega_c|\x)$. 

\newpage 

In case d we see the other extreme, i.e. the expert is overfitted on some data distribution, and thus, the optimization problem becomes essentially the maximization of the expert $p(\x)$ itself.

When using activation maximization for the purpose of model validation, an overfitted expert (case d in Figure~\ref{figure:expert}) must be especially avoided, as the latter could hide interesting failure modes of the model $p(\omega_c|\x)$. A slightly underfitted expert (case b), e.g.\ that simply favors images with natural colors, can already be sufficient. On the other hand, when using AM to gain knowledge on a correctly predicted concept $\omega_c$, the focus should be to prevent underfitting. Indeed, an underfitted expert would expose optima of $p(\omega_c|\x)$ potentially distant from the data, and therefore, the prototype $\x^\star$ would not be truly representative of $\omega_c$.


In certain applications, data density models $p(\x)$ can be hard to learn, or they can be so complex that maximizing them becomes difficult or even intractable. Therefore, a useful alternative class of unsupervised models are \textit{generative models} (e.g., Boltzmann machines, variational Autoencoders, etc.) which do not provide the density function directly, but are able to sample from it, usually via the following two steps:
\begin{enumerate}
\item Sample from a simple distribution $q(\z) \sim \mathcal{N}(0,I)$ which is defined in an abstract code space $\mathcal{Z}$;
\item Apply to the sample a decoding function $g:\mathcal{Z} \to \mathcal{X}$, that maps it back to the original input domain.
\end{enumerate}

One suitable model is the \textit{generative adversarial network (GAN)} introduced by \cite{GoodfellowBengio:2014:adversarialNets}. It consists of two models: a generative model $G$ that captures the data distribution, and a discriminative model $D$ that estimates the probability that a sample came from the training data rather than from $G$. The training procedure for $G$ is to maximize the probability of $D$ making an error - which works like a minimax (minimizing a possible loss for a worst case maximum loss) two-player game. In the space of arbitrary functions $G$ and $D$, a unique solution exists, with $G$ recovering the training data distribution and $D$ equal to $\frac{1}{2}$ everywhere; in the case where $G$ and $D$ are defined by multi-layer perceptrons, the entire system can be trained with backpropagation.

To learn the generator's distribution $p_g$ over data $\bm{x}$, a prior must be defined on the input noise variables $p_{\bm{z}}(\bm{z})$, and then a mapping to the data space as $G(\bm{z}; \theta_g)$, where $G$ is a differentiable function represented by a multi-layer perceptron with parameters $\theta_g$. The second multi-layer perceptron $D(\bm{x}; \theta_d)$ outputs a single scalar. $D(\bm{x})$ represents the probability that $\bm{x}$ came from the data rather than $p_g$. $D$ can be trained to maximize the probability of assigning the correct label to both training examples and samples from $G$. Simultaneously $G$ can be trained to minimize $\log(1-D(G(\bm{z})))$; in other words, $D$ and $G$ play the following two-player minimax game with value function $V(G, D)$:

\begin{equation}
\min_G \max_D V(D, G) = \mathbb{E}_{\bm{x} \sim p_{\text{data}}(\bm{x})}[\log D(\bm{x})] + \mathbb{E}_{\bm{z} \sim p_{\bm{z}}(\bm{z})}[\log (1 - D(G(\bm{z})))].
\end{equation}

\cite{NguyenEtAl:2016:DeepGeneratorNets} proposed to build a prototype for $\omega_c$ by incorporating such a generative model in the activation maximization framework. The optimization problem is redefined as:

\begin{equation}
\max_{\z \in \mathcal{Z}}~ \log p(\omega_c \,|\, g(\z)) - \lambda \|\z\|^2,
\end{equation}

where the first term is a composition of the newly introduced decoder and the original classifier, and where the second term is an $\ell_2$-norm regularizer in the code space. Once a solution $\z^\star$ to the optimization problem is found, the prototype for $\omega_c$ is obtained by decoding the solution, that is, $\x^\star = g(\z^\star)$.

The $\ell_2$-norm regularizer in the input space can be understood in the context of image data as favoring gray-looking images. The effect of the $\ell_2$-norm regularizer in the code space can instead be understood as encouraging codes that have high probability. High probability codes do not necessarily map to high density regions of the input space; for more details please refer to the excellent tutorial given by \cite{MontavonSamekMueller:2017:InterpertingDL}.

\section{Explainable Models for Image Data}
\label{sec:explainable-images}

One technique, which is highly interesting for the medical domain, e.g. for images generated by digital pathology (which are orders of magnitudes larger than e.g. radiological images) \citep{HolzingerEtAl:2017:AugmentedPathologist} is the use of \textit{deconvolutional networks} \citep{ZeilerEtAl:2010:DeconvolutionalNets}. They are an excellent framework that permits the unsupervised construction of hierarchical image representations and thus enables visualization of the layers of convolutional networks \citep{ZeilerFergus:2014:VisUnderstandConvNets}.


Understanding the operation of a convolutional neural network requires the interpretation of feature activity in intermediate layers, and these can be mapped back to the input image space, showing what input pattern originally caused a given activation in the feature maps. This brings us to enormously important fundamental research opportunities in causality \citep{KrynskiTenenbaum:2007:CausalityUncertainty,Pearl:2009:Causality,Bottou:2014MLreasoning} - which is novel in the context of personalized (P4) medicine.

From the perspective of fundamental research, the gained insights might contribute towards building machines that learn and think like people \citep{LakeSalakTenenbaum:2015:ConceptLearning,LakeUlmanTenenbaumGershman2016:MachinesThink}.

One possibility to make e.g. deep networks \citep{LeCunBengioHinton:2015:DeepLearningNature} transparent is to generate image captions in order to train a second network with explanations without explicitly identifying the semantic features of the original network.

Of enormous importance is the possibility to extend the approaches used to generate image captions to train a second deep network to generate explanations \citep{HendricksAkataEtAl:2016:VisualExplanations} – which is at the intersection of images and text and can be tackled with Visual Question Answering (VQA) \citep{GoyalEtAl:2016:VisualQuestionAnswering}. While this second network is not guaranteed to provide reasons correlated to those used in the original network, it seems promising to use neural attention mechanisms to be able to trace which part of the input contributed most to which part of the output, see e.g. \citep{PavlopoulosEtAl:2017:DeeperAttention}.

We envision alternative machine learning techniques that learn more structured, interpretable, and causal models. These can include Bayesian Rule Lists \citep{LethamRudinEtAl:2015:InterpretableClassifiers}, and in order to learn richer, more conceptual and generative models, techniques such as Bayesian Program Learning \citep{LakeSalakTenenbaum:2015:ConceptLearning}, learning models of causal relationships \citep{MaierEtAl:2010:LearningCausalModels,MaierEtAl:2013:LearningCausalModels,AalenEtAl:2016:BelieveDAGs}, and stochastic grammars to learn more interpretable structures \citep{BrendelTodorovic:2011:LearningGraphsHumanActivities,ZhouTorre:2012:FactorizedGraphMatching,ParkNieZhu:2016:AndOrGrammar}. Very useful for building explainable systems in the medical domain is generally genetic programming \citep{Koza:1994:GeneticProgramming,PenaSipper:1999:FuzzyGenetic,TsakonasEtAl:2004:GeneticMedical}, and specifically evolutionary algorithms \citep{WangPalade:2011:InterpretableFuzzy,HolzingerKetAl:2014:DarwinLamarck,HolzingerEtAl:2016:iMLExperiment}.

\section{Explainable Features and Models in Images and *omics data}
\label{sec:explainable-imagesOmics}

\subsection{Image Analysis using Multiscale AM-FM Image Decompositions}
\label{subsec:AMFM-decompositions}

Amplitude Modulation – Frequency Modulation (AM-FM) decompositions provide meaningful representations of medical image and video content. An image is decomposed into a sum of AM-FM components that can be easily visualized in $\mathbb{R}^2$ for humans seeking to understand essential image content.

Input images can be expressed as a sum of AM-FM components, where the challenge is to decompose any input image $s(x)$ into a sum of bi-dimensional AM--FM harmonics of the form

\begin{equation}
s(x_1,x_2)=\sum_{\ell=1}^L s_{\ell}(t) = \sum_{\ell=1}^L A_\ell(x_1,x_2)\cos(\varphi_\ell(x_1,x_2))\;,
\end{equation}

where $A_\ell>0$ denotes a slowly--varying amplitude function, $\varphi_\ell$ denotes the
phase, and $\ell = 1,\cdots,L$ indexes the different AM--FM harmonics. To each phase function, one can associate an instantaneous frequency vector field defined as $\omega_\ell=\nabla\varphi_\ell$. Finding the components $s_\ell$ from the bidimensional signal $s$ is called the {\it decomposition problem}   \citep{ClauselOberlinPerrier:2015:WaveletAMFM}.

\begin{figure}[ht!]
\centering
\includegraphics[width=\textwidth]{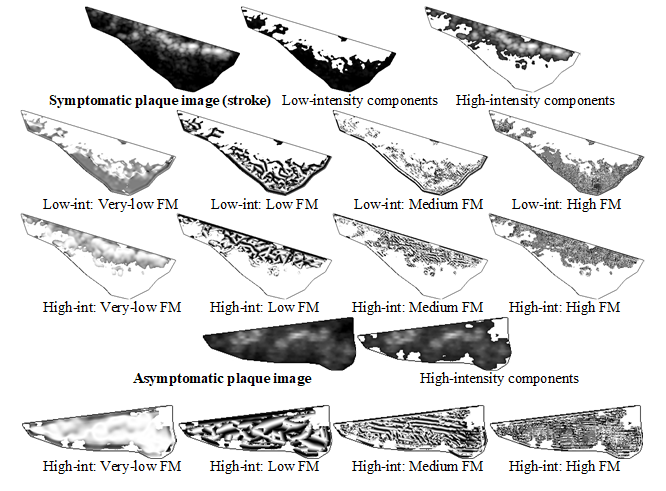}
\caption[Ultrasound Figure]{Multi-scale AM-FM decomposition based on fixed scales (non-adaptive). In the top three rows, we have images from a symptomatic plaque. In the bottom two rows is an asymptomatic example. int = intensity}
\label{fig:fig1}
\end{figure}

We provide an example in Figure \ref{fig:fig1} in a symptomatic stroke plaque and an asymptomatic plaque in ultrasound imaging of the carotid. In top three rows of Figure  \ref{fig:fig1}, we have a symptomatic example that can be used to demonstrate several issues associated with high-risk cases. First, we have large dark regions corresponding to the lipids or other dangerous components. Second, these dark plaque regions are located very close to the plaque surface. However, in the original images, we cannot see any structure over these dark regions. A very rich structure plaque surface structure is revealed by the FM reconstructions of the second row. Starting from the very-low to the high frequency scales, instantaneous frequency  can be seen adjusting to the local texture content with some sharp changes around different structures.
In contrast, the asymptomatic plaque image reconstructions of the last two rows do not include significant low-intensity components. The high-intensity components of the fourth row (right image) dominate the reconstruction.
There is also more regularity (homogeneity) in the asymptomatic reconstructions of the last row. Far more variability and heterogeneity are evident in the symptomatic FM reconstructions.

AM-FM decompositions have been enabled following the introduction of new demodulation methods as summarized in \citep{MurrayEtAl:2010:MultiscaleAMFM}, \citep{MurrayEtAl:2012:MultiscaleImaging} and are a hot topic, and in a combined effort can be very beneficial for context understanding; a summary of several medical applications of novel multi-scale AM-FM methods can be found in our recently published survey \citep{MurrayEtAl:2012:MultiscaleImaging}. Earlier work with AM-FM models had demonstrated its promise with textured images, as in the example of fingerprint image classification \citep{PattichisEtAl:2001:FingerprintAMFM}, tree image analysis analysing growth seasons \citep{RamachandranEtAl:2011:TreeImageGrowth}, non-stationary wood-grain characterization, and other texture images \citep{KokkinosEtAl:2009:TextureAnalysis}.


The introduction of a multiscale approach in \citep{MurrayEtAl:2010:MultiscaleAMFM} demonstrated that the method can be used to reconstruct general images. In particular, a multi-scale AM-FM representation led to the important application of population screening for diabetic retinopathy\footnote{Early detection of diabetic retinopathy is extremely important in order to prevent premature visual loss and blindness} as documented in \citep{AgurtoEtAl:2010:MultiscaleDiabeticRetinopathy}, \citep{RahimEtAl:2015:Retinopathy}, hysteroscopy image assessment \citep{ConstantinouEtAl:2012:AdaptiveMultiscale}, fMRI and MRI image analysis \citep{LoizouEtAl:2011:MultiscaleMRI}, and atherosclerotic plaque ultrasound image and video analysis \citep{LoizouEtAl:2011:MultiscaleCarotid}. Alternatively, the definition of multidimensional AM-FM transforms over curvilinear coordinate systems has led to the earlier development of very low bitrate video coding as demonstrated in \citep{LeePattichisBovik:2001:FoveatedVideo,LeePattichisBovik:2002:FoveatedQuality}.

Complex wavelets can also be very powerful, e.g. in the analysis of images of electrophoretic gels used in the analysis of protein expression levels in living cells, where much of the positional information of a data feature is carried in the phase of a complex transform. Complex transforms allow explicit specification of the phase, and hence of the position of features in the image. Registration of a test gel to a reference gel is achieved by using a multiresolution movement map derived from the phase of a complex wavelet transform (the Q-shift wavelet transform) to dictate the warping directly via movement of the nodes of a Delaunay-triangulated mesh of points. This warping map is then applied to the original untransformed image such that the absolute magnitude of the spots remains unchanged. The technique is general to any type of image. Results are presented for a simple computer simulated gel, a simple real gel registration between similar “clean” gels with local warping vectors distributed along one main direction, a hard problem between a reference gel and a “dirty” test gel with multi-directional warping vectors and many artifacts, and some typical gels of present interest in post-genomic biology. The method compares favourably with others, since it is computationally rapid, effective and entirely automatic \citep{WoodwardRowlandKell:2004:proteomeImages}.

\subsection{Meaningful AM-FM Decompositions and Features in Artificial Neural Networks}
\label{am-fm-ANN}

A key purpose of our proposal for explainable methods is to extend current deep learning methods to incorporate meaningful AM-FM decompositions and features into the classification models. Here, we propose to investigate the use of AM-FM based Artificial Neural Networks (AM-FM ANNs), where two separate approaches remain open for investigation: (i) Multiscale AM-FM ANNs, and (ii) Hybrid AM-FM ANN architectures. In our first system, we will investigate the use of AM-FM derived features in an extended ANN. Here, our goal is to extract multiscale feature maps based on the instantaneous amplitude and the instantaneous frequency that will be fed as inputs to a multilayer neural network for medical image classification purposes. In our second system, we want to develop a hybrid system by incorporating AM-FM decomposition in a hybrid ResNet architecture. Here, we are motivated by the fact that ResNet models have provided excellent classification results through the use of convolutional neural networks \citep{JegouEtAl:2012:AggregatingLocalImage,XieEtAl:2016:AggregatedResidualDeep,HuangPeng:2017:DeepMetric,Taki:2017:DeepResidualNetworks}.

The lower layers of the ResNet architecture make heavy use of convolutional layers to represent the input. ResNet relies on the use of skip connections to compute residuals resulting from the use of a group of convolutional layers.

Unfortunately, the use of 152 layers makes it very difficult to understand the internal structure of such networks. Consequently, we propose a hybrid system, where our goal is to use AM-FM decompositions to replace the overwhelming majority of the lower-level layers. The use of multiple AM-FM components will lead to a significant reduction in the residual representation of the input image. The significantly reduced residuals will be incorporated into a ResNet architecture that will focus on training a small number of upper-level layers. Overall, we will visualize all convolution layers through their frequency coverage. In the frequency domain, each filter will be characterized by an effective 2D spatial frequency band that will capture image periodicities in specific directions.

\begin{figure}[ht!]
\centering
\includegraphics[width=\textwidth]{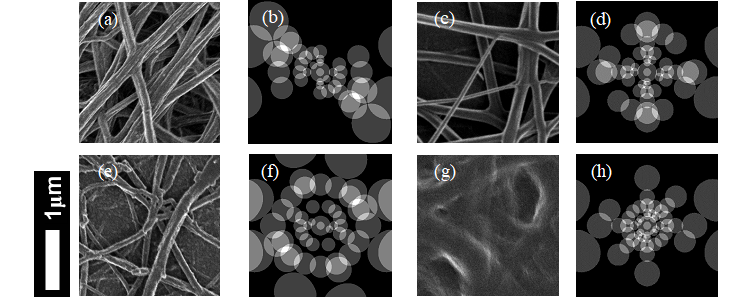}
\caption[SEM Figure]{SEM (Scanning Electron Microscopy) images of diabetes plasma with corresponding dominant Gabor filters. (a) and (c) Original nice spaghetti-like images in healthy controls. (b) and (d) Dominant Gabor filters for (a) and (c) respectively. (g) Dense matted deposits (DMDs) in type 2 diabetes that are removed (e) when we add small amounts of lipopolysaccharide-binding protein (LBP). (f) and (h) Dominant Gabor filters for (e) and (g) respectively. The dominant Gabor filters are shown in the frequency domain. SEM images taken from 2017 Nature Scientific Reports Diabetes and Control Data, Figure 7A.}
\label{fig:fig2}
\end{figure}

Co-author Kell has been developing novel methods to assess the severity of chronic, inflammatory diseases from the morphology of blood clots seen in either the Scanning Electron Microscope (SEM) or – when amyloid-specific stains are added – by confocal microscopy, see e.g. \cite{KellPretorius:2017:Proteins}. As yet, the analyses of the images have been either very simple or qualitative. We present four examples of AM-FM decompositions of such SEM images in Figure \ref{fig:fig2}, where we show the original input images and the corresponding dominant Gabor filters.

In our example, we compute an AM-FM component from each Gabor filter. Then, the dominant Gabor filters are determined by requiring an excellent reconstruction that satisfies (Structural SIMilarity Index (SSIM) $>$ 0.85, not shown in Figure \ref{fig:fig2}). In Figure \ref{fig:fig2}, we show the frequency magnitude plots of the dominant Gabor filters. Overlapping regions appear brighter.

From Figure \ref{fig:fig2}, it is clear that the list of the dominant Gabor filters provides for a very compact visualization of image content. In Figure \ref{fig:fig2}, each symmetric pair of circles represents a single filter. With just 10 to 30 filters, we can describe strong variabilities in image content. The frequency domain is also very easy to explain. We observe strong directional selectivity orthogonal to image lines, strong concentration of low-frequency components, and select, high frequency components. As described earlier, these decompositions have provided excellent features for a wide-range of biomedical applications. Furthermore, in comparison, ResNet requires 152 layers that cannot be easily visualized.

\subsection{*omics spectrum and Bioinformatics}
\label{omicsspectrum}

Large-scale *omics are important footprints of biological mechanisms. These mechanisms consist of numerous synergistic effects emerging from various systems of interwoven biomolecules, cells and tissues.  Modern biology harnesses its power from technological advances in the field of *omics and the advent of next generation sequencing (NGS). These technologies provide a spectrum of information ranging from genomics, transcriptomics and proteomics to epigenomics, pharmacogenomics, metagenomics and metabolomics, etc. The sheer size of data generated by these high-throughput methodologies necessitates the need to analyse, integrate and concurrently interpret this avalanche of information in a systemic way.

Currently, platforms are missing that help towards not only the analysis but the \textit{interpretation} of the data, information and knowledge obtained from the above-mentioned omics technologies and to cross-link them to hospital data. Moreover, it is necessary to narrow down the gap between genotype and phenotype as well as providing additional information regarding biomarker research and drug discovery, where biobanks \citep{Huppertz:2014:Biobank} play an increasingly important role.

One of the grand challenges here is to close the research cycle in such a way that all the data generated by one research study can be consistently associated with the original samples, therefore the underlying original research data and the knowledge gained thereof, can be reused. This can be enabled by a catalogue providing the information hub connecting all relevant information sources \citep{MuellerHolzinger:2015:BiobankIntegration}. The key knowledge embedded in such a biobank catalogue is the availability and quality of proper samples to perform a research project. To overview and compare collections from different catalogues, visual analytics techniques are necessary, especially glyph based visualization techniques \citep{MuellerEtAl:2014:MultilevelGlyphs}. We cannot emphasize often enough the \textbf{combined view on heterogeneous data sources in a unified and meaningful way,} consequently enabling the discovery and visualization of data from different sources, which would enable totally new insights.

Here, toolsets are urgently needed to support the bidirectional interaction with computational multiscale analysis and modelling to help to go towards the far-off goal of future medicine\footnote{often called P4-medicine, i.e. medicine that is Personal, Participatory, Predictive and Preventive} \citep{HoodFriend2011P4cancermedicine,TianPriceHood:2012:P4medicine}.

\section{Explainable Models for Text}
\label{explainbable-text}

Text is inherently different from continuous data such as images since it is composed of (natural language) symbols used for communication between humans. The formalization of natural language has been a long-standing effort of philosophy, starting from the Platonic cave of ideas and culminating in the vision of the Semantic Web \citep{bernerslee2001semantic}.

However, while logical reasoning with traces of inference steps (see above) is rather well-understood, automated text understanding is still hampered by the fact that text is more expressive than tractable forms of logic and their representations, e.g. see \citep{Krotzsch2008} and there are no established ways to convert a text to its ontological representation \citep{Cimiano14}. 

Ontology learning generally \citep{MaedcheStaab:2001:OntologyLearning}, and ontology learning from text specifically is a hot topic \citep{CimianoVoelker:2005:Text2Onto}. Ontology reasoning, nowadays based on deep learning rather than logic-based formal reasoning \citep{HoheneckerLukasiewicz:2017:DeepLearningOntology} is of great interest, and the applications of ontology-guided approaches are of much practical help for the medical domain expert \citep{Girardi:2016:iKDDdocInLoop,WartnerEtAl:2016:Limits-Doc-in-Loop}.

Generally, in the field of natural language processing, symbolic methods, which are inherently more interpretable, have never fallen out of use since their heydays in the 1970ies, but have been mostly replaced with statistical, probabilistic and more recently deep neural network models. 

On the side of interpretability, this meant going from brittle rule-based but glass-box approaches to more robust opaque-box approaches, with supporting statistics but no explicit 'real reasons', to ultimately neural black-box approaches. 

In such black-box approaches already the input symbols (words) are replaced by vectors (e.g. skip-gram model for learning vector representations of words from large amounts of unstructured text \citep{MikolovChenCorradoDean:2013:WordVector}), resulting in a few hundred un-interpretable dimensions (a.k.a. embeddings, e.g. \citep{MikolovEtAlDean:2013:RepresentationsWords}).

As opposed to fields such as speech or image processing, the improvements recently gained with deep learning on text are rather modest, yet its use is very attractive since neural representations reduce the workload of manually crafting features enormously \citep{Manning2015}.

In the medical domain, where a large amount of knowledge is represented in textual form, there exists already a large knowledge graph of medical terms (the UMLS\footnote{https://www.nlm.nih.gov/research/umls}), where it is crucial to underpin machine output with reasons that are human-verifiable and where high precision is imperative for supporting, not distracting practitioners. The only way forward seems to be the integration of both knowledge-based and neural approaches to combine the interpretability of the former with the high efficiency of the latter.
To this end, there have been attempts to retrofit neural embeddings with information from knowledge bases (e.g. \citep{Faruqui15}) as well as to project embedding dimensions onto interpretable low-dimensional sub-spaces \citep{rothe-ebert-schutze:2016:N16-1}.

More promising, in our opinion, is the use of \textit{hybrid distributional models} that combine sparse graph-based representations \citep{BiemannRiedl13} with dense vector representations \cite{MikolovEtAlDean:2013:RepresentationsWords} and link them to lexical resources and knowledge bases \citep{Faralli16}. Here a \textbf{hybrid human-in-the-loop approach} can be beneficial, where not only the machine learning models for knowledge extraction are supported and improved over time, the final entity graph becomes larger, cleaner, more precise and thus more usable for domain experts \citep{YimamEtAl:2017:BioNLP}. Contrary to classical automatic machine learning, human-in-the-loop approaches do not operate on predefined training or test sets, but assume that human expert input regarding system improvement is supplied iteratively. In such an approach the machine learning model is built continuously on previous annotations and used to propose labels for subsequent annotation \cite{YimamEtAl:2016:AdaptiveAnnotation}.

\begin{figure}[ht!]
\centering
\includegraphics[width=\textwidth]{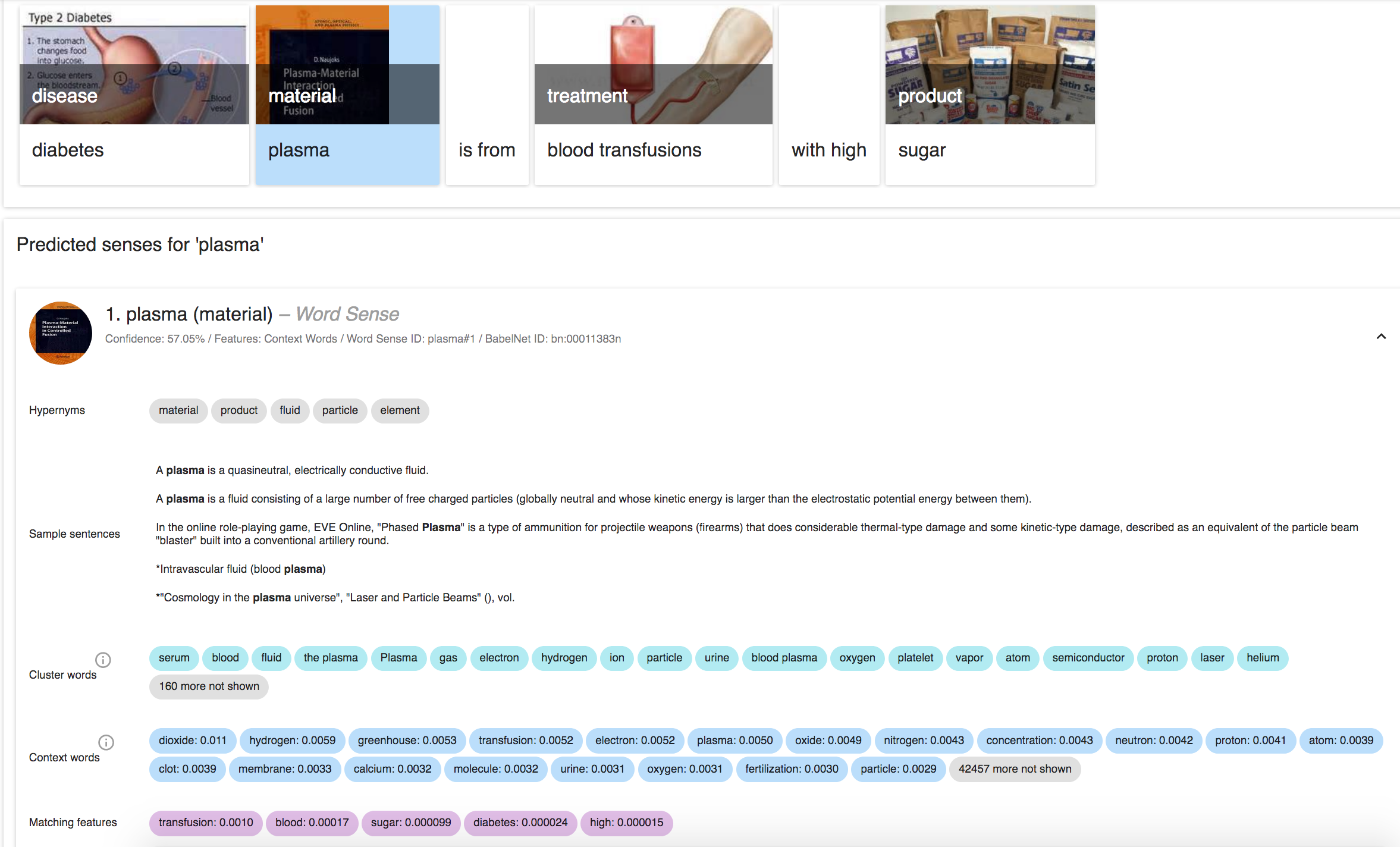}
\caption{Output of an unsupervised interpretable model for text interpretation for the input: "diabetes plasma is from blood transfusions with high sugar" (Note that it brings up plasma (material) not plasma (display device)! Image created online via ltbev.informatik.uni-hamburg.de/wsd on 27.12.2017, 19:30}
\label{fig:plasma}
\end{figure}

Combined with an interpretable disambiguation system \citep{panchenko-EtAl:2017:EACLlong}, this realizes concept-linking in context with high accuracy while providing human-interpretable reasons for \textit{why} concepts have been selected.
Figure \ref{fig:plasma} shows machine reading capabilities of the system described in \cite{panchenko-EtAl:2017:EACLlong}: The system can automatically assign more general terms in context and can disambiguate terms with several senses to the one matching the context. Note that while unsupervised machine learning is used for inducing the sense inventory, the senses are interpretable by providing a rich sense representation as visible in the figure. This method does not require a manually defined ontology and thus is applicable across languages and domains.

The quest for the future is to generalize these notions to enable semantic matching beyond keyword representations (cf. \cite{cer-EtAl:2017:SemEval}) in order to transfer these principles from the concept level to the event level.

\section{Conclusion and Future Outlook}

Explainable-AI systems generally pose huge challenges yet offer enormous opportunities for the medical domain. Recent research demonstrates examples of promising directions, but to date none of these examples provides a complete solution, nor are these considered the only possible solutions.

If human intelligence is complemented by machine learning and at least in some cases even overruled, humans must be able to understand, to re-enact and to be able to interactively influence the machine decision process \cite{Holzinger:2016:iML}. The European Union cannot afford to deploy highly intelligent AI systems that make unexpected decisions which cannot be reproduced, especially if these systems are in the medical domain. Consequently, a huge motivation for explainable-AI are rising legal and privacy aspects. The new European General Data Protection Regulation (GDPR 2016/679 and ISO/IEC 27001) entering into force on May, 25, 2018, will make black-box approaches difficult to use in business, if they are not able (on demand) to explain \textit{why} a decision has been made. "Potentially interpretable" could also mean "accurate", see for example the recent work "One pixel attack for fooling deep neural networks" by \cite{SuEtAl:2017:FoolingDeep}. Also small changes in images may lead to huge changes in the prediction in deep learning approaches, see "Deep Neural Networks are Easily Fooled" by \cite{NguyenYosinskiClune:2015:DeepFooled}. None of this suggests easy interpretability is even possible for deep neural networks.

In the medical domain a large amount of knowledge is represented in textual form, and the written text of the medical reports is legally binding, unlike images nor *omics data. Here it is crucial to underpin machine output with reasons that are human-verifiable and where high precision is imperative for supporting, not distracting the medical experts. The only way forward seems the integration of both knowledge-based and neural approaches to combine the interpretability of the former with the high efficiency of the latter. Promising for explainable-AI in the medical domain seems to be the use of \textbf{hybrid distributional models} that combine sparse graph-based representations with dense vector representations and link them to lexical resources and knowledge bases.

Last but not least we emphasize that successful explainable-AI systems need effective user interfaces, fostering new strategies for presenting \textbf{human understandable explanations,} e.g. explanatory debugging \citep{KuleszaEtAl:2015:ExplanatoryDebugging}, affective computing \citep{Picard:1997:AffectiveComputingBook}, sentiment analysis \citep{MaasNgEtAl:2011:WordVectorsSentiment,PetzEtAl:2015:Sentiment}. While the aforementioned methods are inherently more explainable, their performance is less optimal hence possibilities to enhance 2-way interaction have to be explored, which calls for optical computing for machine learning purposes. 

\subsubsection*{Acknowledgments}

We are grateful for valuable comments from our local, national and international colleagues, including George Spyrou, Cyprus Institute of Neurology and Genetics, Chris Christodoulou, Ioannis Constantinou and Kyriacos Constantinou, University of Cyprus and Marios S. Pattichis, University of New Mexico.

\bibliographystyle{unsrtnat}
\bibliography{references}

\end{document}